\documentclass[letterpaper]{article} 
\usepackage{aaai23}  
\usepackage{times}  
\usepackage{helvet}  
\usepackage{courier}  
\usepackage[hyphens]{url}  
\usepackage{graphicx} 
\usepackage{natbib}  
\usepackage{caption} 
\usepackage[T1]{fontenc}

\usepackage{algorithm}
\usepackage{algorithmic}
\usepackage{multirow}
\usepackage{amssymb}
\usepackage{color}
\usepackage{rotating}
\usepackage{xcolor}
\usepackage{colortbl}
\usepackage{amsmath}
\usepackage{booktabs}
\usepackage{array}
\usepackage{newfloat}
\usepackage{listings}
\DeclareCaptionStyle{ruled}{labelfont=normalfont,labelsep=colon,strut=off} 
\floatstyle{ruled}
\newfloat{listing}{tb}{lst}{}
\floatname{listing}{Listing}

\nocopyright

\title{AdvMono3D: Advanced Monocular 3D Object Detection \\
		with Depth-Aware Robust Adversarial Training}
\author{
	Xingyuan Li\textsuperscript{\rm 1},
	Jinyuan Liu\textsuperscript{\rm 1},
	Long Ma\textsuperscript{\rm 1},
	Xin Fan\textsuperscript{\rm 1},
	Risheng Liu\textsuperscript{\rm 1,2\thanks{Corresponding author}}
}
\affiliations{
	\textsuperscript{\rm 1}Dalian University of Technology, China\\
	\textsuperscript{\rm 2}Peng Cheng Laboratory, China\\
	xingyuan\_lxy@163.com, atlantis918@hotmail.com, malone94319@gmail.com, \{xin.fan, rsliu\}@dlut.edu.cn
}
\usepackage{bibentry}

\begin{document}

\maketitle

\begin{abstract}
    Monocular 3D object detection plays a pivotal role in the field of autonomous driving and numerous deep learning-based methods have made significant breakthroughs in this area. Despite the advancements in detection accuracy and efficiency, these models tend to fail when faced with such attacks, rendering them ineffective. Therefore, bolstering the adversarial robustness of 3D detection models has become a crucial issue that demands immediate attention and innovative solutions. To mitigate this issue, we propose a depth-aware robust adversarial training method for monocular 3D object detection, dubbed DART3D. Specifically, we first design an adversarial attack that iteratively degrades the 2D and 3D perception capabilities of 3D object detection models(IDP), serves as the foundation for our subsequent defense mechanism. In response to this attack, we propose an uncertainty-based residual learning method for adversarial training. Our adversarial training approach capitalizes on the inherent uncertainty, enabling the model to significantly improve its robustness against adversarial attacks. We conducted extensive experiments on the KITTI 3D datasets, demonstrating that DART3D surpasses direct adversarial training (the most popular approach) under attacks in 3D object detection $AP_{R40}$ of car category for the Easy, Moderate, and Hard settings, with improvements of 4.415\%, 4.112\%, and 3.195\%, respectively.
\end{abstract}

\section{Introduction}

3D object detection is of paramount importance in various applications, including autonomous driving and robotics, and has attracted significant research attention in recent years. Substantial progress has been made, particularly for LiDAR and stereo-based approaches \cite{zhou2018voxelnet,xu2021spg,sheng2022rethinking,yang2022graph}. However, monocular 3D object detection\cite{wang2021depth,manhardt2019roi,chen2016monocular} remains an attractive alternative due to its lower cost and simpler configuration, making it more accessible for a wide range of applications. 

\begin{figure}[ht]
	\centering
	\includegraphics[scale=0.49]{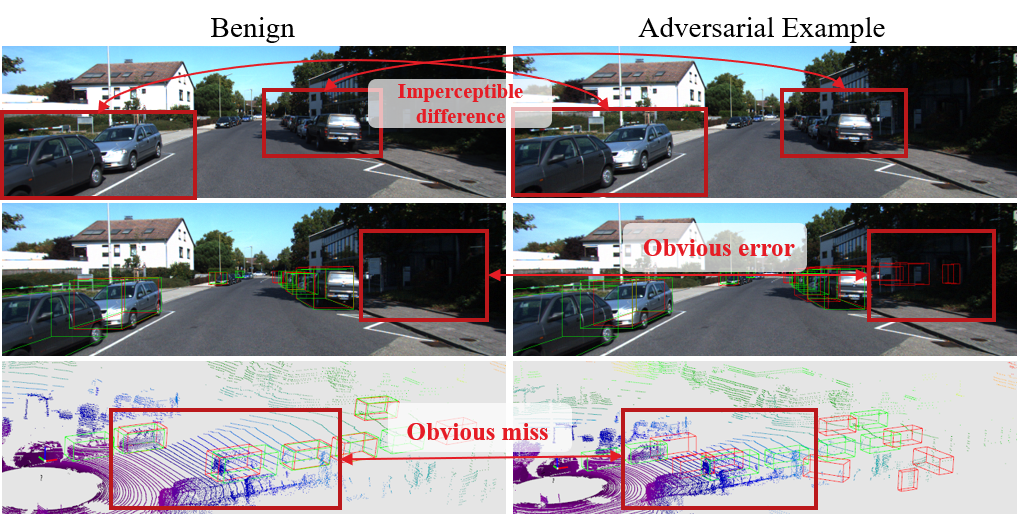}
    \caption{Adversarial attack demonstration on GUPNet. \textcolor{green}{Green boxes}: ground truth; \textcolor{red}{Red boxes}: detection results. Top row shows input images from KITTI 3D benchmark and adversarial examples generated by IDP. Middle and Bottom row draws regular and adversarial attack-affected 3D object detection results on image plane and LiDAR point cloud images, respectively.}
\end{figure}

In recent years, monocular 3D object detection has experienced significant progress, with researchers exploring various strategies to address the inherent challenges associated with depth estimation and object localization. These approaches can be classified into several categories based on the techniques they employ to tackle the depth estimation problem. Direct regression approaches attempt\cite{chen2016monocular} to estimate instance depth by regressing the depth values directly from the input image. Geometry-based methods\cite{brazil2019m3d} have been proposed to leverage the geometric relationships within the scene to aid in depth estimation. Error-aware techniques\cite{ma2021delving} focus on analyzing the errors in depth estimation and developing strategies to mitigate these errors. 

Although monocular 3D object detection has achieved numerous breakthroughs, its robustness against adversarial attacks remains an open challenge. The lack of robustness can lead to several issues, including inaccurate depth estimation, erroneous object localization, and distortion of 3D information, which in turn could significantly impact the performance of applications reliant on monocular 3D detection. The primary reason for this vulnerability is the absence of comprehensive analyses and defense strategies specifically tailored for 3D object detection under adversarial attacks. While some works \cite{zhang2019theoretically,shafahi2019adversarial} have introduced adversarial training strategies to enhance the robustness of deep learning models, these methods may not be directly applicable to monocular 3D object detection owing that it relies heavily on a single image, which inherently lacks depth information. 

To address the aforementioned challenges, we introduce a novel adversarial training approach, Depth-aware Robust Adversarial Training for Monocular 3D Object Detection (DART3D). Initially, we devise an innovative adversarial attack strategy that iteratively targets the 2D and 3D perception capabilities of the image. This targeted approach significantly degrades the performance of 3D detection compared to other attacks such as FGSM\cite{goodfellow2014explaining}, BIM\cite{kurakin2016adversarial}, PGD\cite{madry2017towards}, and MI-FGSM\cite{dong2018boosting}. Building upon this attack, we propose a residual learning method grounded in uncertainty. This method incorporates a pseudo-label which is generated using a method that is opposite to the generation of adversarial noise, enabling the model to implicitly understand the depth information contained within the noise while learning about the noise itself. The element of uncertainty is introduced through a concealment mechanism embedded within our model. This mechanism randomly activates the denoising module, creating an unpredictable environment for the attacker. The inherent uncertainty in whether the denoising module will be activated enhances the model's defense capabilities against adversarial attacks, making DART3D a robust solution for monocular 3D object detection. 

We showcase the effectiveness of our method through experiments on the KITTI dataset and adversarial attack samples, where it exhibits superior performance compared to state-of-the-art methods and successfully defends against diverse adversarial attacks. In summary, our contributions are as follows:

\begin{itemize}
    \item
    We propose an adversarial attack specifically designed for monocular 3D object detection, which effectively disrupts the depth perception of 3D detectors. This is the first work to introduce such a targeted attack in the field of monocular 3D object detection.
    \item
    We introduce an adversarial training strategy that leverages uncertainty to learn from noise information. This innovative approach effectively mitigates the impact of adversarial attacks and enhances the robustness of the model.
    \item
    Our adversarially robust 3D detection model outperforms other 3D detection methods and adversarial training strategies, achieving state-of-the-art performance. This demonstrates the effectiveness and superiority of our proposed method in the context of adversarial robustness for 3D object detection.
\end{itemize}

\begin{figure*}[ht]
	\centering
	\includegraphics[scale=0.63]{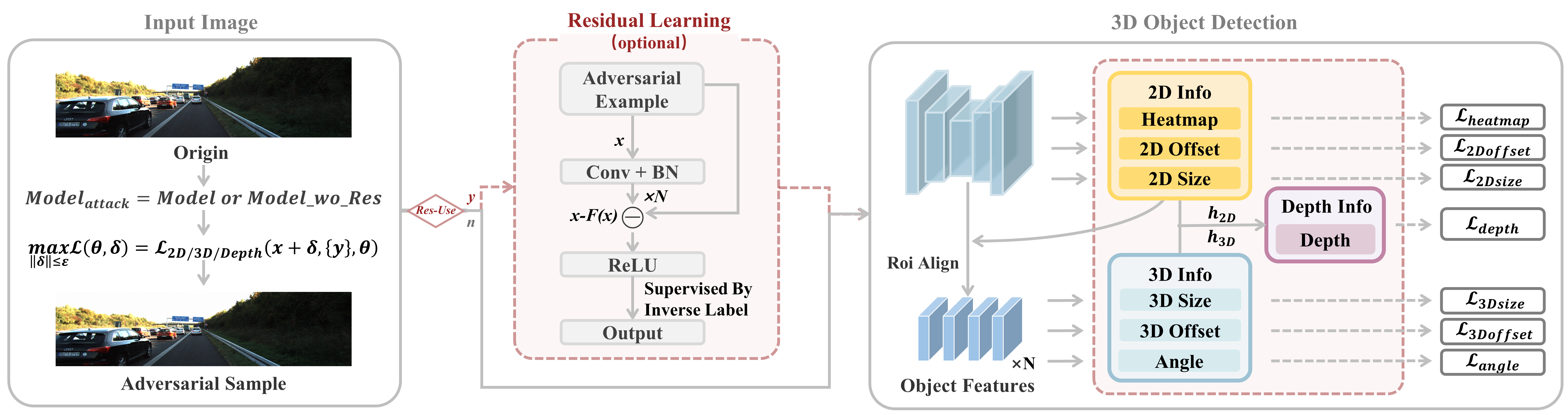}
	\caption{The framework of our proposed method. First, the attacker chooses the method for generating adversarial noise, followed by the defender's decision to activate the Residual Learning or not. Subsequently, the input is processed by the 3D detection network, which regresses 2D information, 3D information, and depth information.}
\end{figure*}

\section{Related works}
\subsection{Adversarial Training}
Adversarial attacks are a type of assault that can disrupt various tasks by introducing subtle noise into images. Adversarial attack has been extensively studied in classification models \cite{goodfellow2014explaining,madry2017towards,liu2022attention}. As learning-based models\cite{liu2012fixed,wu2019essential,liu2021learning,liu2022target,liu2023multi}, deep networks\cite{piao2019depth,piao2020a2dele,zhang2020select,liu2021retinex,liu2020real,ma2023bilevel,ma2022practical,liu2020bilevel} are vulnerable to adversarial examples' influence \cite{szegedy2013intriguing}. Adversarial training is an effective defense method for countering such attacks. The two representative methods for generating white-box adversarial attacks are the Fast Gradient Sign Method (FGSM) \cite{goodfellow2014explaining} and Projective Gradient Descent (PGD) \cite{madry2017towards}. Recently, numerous adversarial attacks have been developed against object adversarial models. Xie et al. \cite{xie2017adversarial} proposed an attack called DAG, which specifies adversarial labels and uses backpropagation to mislead predictions. Li et al. \cite{li2018robust} employed label loss and shape loss to generate adversarial perturbations. Wei et al. \cite{wei2018transferable} utilized the Generative Adversarial Network (GAN) framework to generate adversarial examples. \cite{wiyatno2018maximal} reformulated the original optimization problem by conducting universal attacks on multiple instances across different classes simultaneously as shown in following equations:
\begin{equation}
\max _\delta \mathcal{L}(\theta, \delta)=\frac{1}{N} \sum_{i=1}^N l\left(x_i+\delta,\left\{y_i\right\}, \theta\right) { s.t. }\|\delta\|_p \leq \epsilon,
\end{equation}

\begin{equation}
\min _\theta \max _{\|\delta\|_p \leq \epsilon} \mathcal{L}(\theta, \delta)=\frac{1}{N} \sum_{i=1}^N l\left(x_i+\delta,\left\{y_i\right\}, \theta\right),
\end{equation}
 
\noindent where $\delta$ is the adversarial perturbation, and $\|\delta\|_p \leq \epsilon $ denotes the ${\mathcal l}_{p}-norm$ to prevent $\delta$ from growing too large.

\subsection{3D Object Detection}
Monocular 3D object detection has gained popularity due to its low cost and simple setup. Existing methods for monocular 3D object detection primarily focus on two aspects: deep representation learning\cite{roddick2018orthographic} and geometric priors \cite{wang2019pseudo}. Pioneering methods \cite{chen2016monocular} integrate different information, such as segmentation and scene prior, for 3D detection. Deep3DBox \cite{mousavian20173d} regresses object dimensions and orientations and employs geometric priors to enforce constraints between 2D and 3D boxes. M3D-RPN \cite{brazil2019m3d} proposed depth-aware convolution for monocular 3D prediction, resulting in improved 3D region proposals. MonoPair \cite{chen2020monopair} parsed spatial pairwise relationships of objects to enhance monocular 3D detection performance. GUPNet \cite{lu2021geometry} tackled the issue of error amplification through geometric guided depth uncertainty. Some studies use depth information to assist in 3D object detection \cite{ma2019accurate,wang2021depth}. CaDDN \cite{reading2021categorical} employs aerial view (BEV) to represent the classification depth distribution of each pixel. OFTNet \cite{roddick2018orthographic} projects 2D image features into 3D space. RoI-10D \cite{manhardt2019roi} utilized CAD models to increase training samples with the aid of depth estimation maps.

\section{Proposed Method}
While current 3D detection models have achieved significant advancements, their resilience against adversarial perturbations remains largely untested. To specifically target the 3D detectors' depth perception, we introduce the Iterative Deterioration of Perception (IDP) attack method. This technique iteratively impairs the 3D detection model's grasp of 2D, 3D and depth information. Alongside this, we craft an adversarial training strategy rooted in the principles of IDP, as depicted in Figure 2. Within our adversarial training framework, we embed an element of uncertainty, allowing the model to interpret adversarial noise with a defined probability. This integration further fortifies the model's defense mechanisms against potential threats.

\subsection{Attack Framework}
The goal of the adversarial attack on monocular 3D detection models is to inject a subtle perturbation into the given input image, such that the perturbation is not visually perceptible, yet leads to a significant degradation in the detector's perception of the object's size, position, and depth. To achieve this, we develop an algorithm based on the idea of MI-FGSM\cite{dong2018boosting}, which is one of the most widely used strong attacks for classification models.

Unlike classification tasks that typically involve a single loss, 3D detection tasks are often more complex. Generally, the 3D detection system consists of three primary components: the 2D estimation, involving the assessment of an object's 2D size and position; the 3D estimation, detailing the object's 3D dimensions and position offset; and the depth estimation of the object, which is often the most intricate aspect. To more effectively disrupt the 3D detection system with adversarial attacks, we utilize losses from heterogeneous sources that stem from different classification and regression tasks when generating adversarial examples. We can formulate an optimization problem to construct an adversarial examples generation approach specifically tailored for 3D detection, as follows:

\begin{equation}
	\begin{aligned} 
		&\max _{\|\delta\|_p \leq \epsilon} \mathcal{L}(\theta, \delta)  =\ell_{2d}(x+\delta,\{y_{2d}\}, \theta) + \\ 
		&\ell_{3d}(x+\delta,\{y_{3d}\}, \theta) + \ell_{depth}(x+\delta,\{y_{depth}\}, \theta),
	\end{aligned}
\end{equation}	

\noindent where $\hat{l}_{t \in{2d, 3d, depth}}(\cdot,{y_{O_{2d}}, y_{S_{2d}}}, \theta)$ denotes the 2D offset loss and localization loss employed for monocular 3D detectors, respectively. Given that 3D detection models prioritize mastering 2D information before leveraging it to estimate 3D attributes, and subsequently integrate both 2D and 3D insights to determine object depth, our attack framework modifies Equation 3 to implement an iterative assault on 2D, 3D, and depth information, aiming for a more potent adversarial impact, as shown in Algorithm 1. Then we can get final adversarial example $x’ = x + \delta$

\begin{algorithm}[t]
	\caption{IDP Attack}
	\begin{algorithmic}[1]
	\REQUIRE image $x$, label $y$, perturbation bound $\epsilon$,  step size $\beta$, network weights $\theta$
    \STATE Sample random noise $\delta$
	\FOR {$iter$ = 1 to $m$}
	\IF{$iter \equiv 1 \;(mod \;3)$}
	\STATE $d_\delta \gets \nabla_{\delta} \hat{l}_{2d}((x+\delta),\{y_{2d}\}, \theta)$
	\ELSIF{$iter \equiv 2 \;(mod \;3)$}
	\STATE $g_\delta \gets \nabla_{\delta} \hat{l}_{3d}((x+\delta),\{y_{3d}\}, \theta)$
	\ELSE
	\STATE $g_\delta \gets \nabla_{\delta} \hat{l}_{depth}((x+\delta),\{y_{depth}\}, \theta)$
	\ENDIF
	\STATE $\delta \gets \delta + \beta \cdot \textit{sign}(g_\delta)$
	\STATE $\delta \gets \{\delta \: | \: \|\delta\|_{\infty} \leq \epsilon\}$
	\ENDFOR
	\RETURN $\delta$
	\end{algorithmic}
\end{algorithm}

\subsection{Uncertainty-Based Residual Learning}
In this section, we introduce a uncertainty-based residual learning methodology, specifically designed to extract adversarial perturbations from adversarial instances. At its core, adversarial noise can be perceived as a unique type of noise, predominantly induced by external security challenges. As adversarial examples are crafted, this noise seamlessly integrates and carries significant label information, further emphasizing the intricate nature of adversarial attacks.

Assuming that the residual mapping is considerably more direct to learn compared to the original unreferenced mapping, the residual learning strategy enables deep CNNs to effectively learn the residual mapping for a few stacked layers. To facilitate this process, we employ the following loss function:

\begin{equation}
	\ell_{r} = ||\hat{x}' - \hat{x}||_{2},
\end{equation}

\noindent where $\hat{x}'$ denote the output of the residual learning given $x'$ as input and $\hat{x}$ is the corresponding label which is formulated as the image subtracted by the adversarial noise.. In the context of adversarial noise, the residual learning formulation is particularly beneficial, as it allows the model to efficiently capture and represent the complex patterns of adversarial perturbations. This, in turn, empowers the model to discern label-related information within adversarial samples and achieve improved detection performance through residual learning.

An essential feature of our approach is the uncertainty-based residual denoising module. Given the minimal impact of slight noise on the image features and geometrical consistency, our training regimen can randomly decide to employ or forego this residual denoiser. The stochastic nature of its application during training provides robustness against potential adversarial attacks, as attackers cannot deterministically predict its employment. Similarly, during testing, the model also has the flexibility to utilize the denoiser randomly. This unpredictability challenges adversaries and results in enhanced protection against adversarial attempts. The entire training strategy is elucidated in Algorithm 2.

\begin{algorithm}[t]
	\caption{Uncertainty-based Adversarial Training}
	\begin{algorithmic}[1]
	\REQUIRE dataset $D$, label $y$, training epoch $T$, learning rate $\gamma$, network weights $\theta$
	\FOR{epoch = 1, ..., $T$}
	\FOR{minibatch B $\sim$ D}
        \STATE \% Adversarial Attack
        \STATE  Generate $\delta$ as Algorithm1
		\STATE $x' \gets x + \delta$
        \STATE \% Adversarial Training
		\STATE  Generate inverse label $\hat{x}$
		\STATE $\hat{x} \gets x - \delta$
		\STATE  Introduce Uncertainty
		\STATE $r \sim \mathcal{U}(0,1)$
		\IF {$r < 0.5$}
		\STATE $\theta \gets \theta - \gamma\mathbb{E}_{x \in B}[\nabla_\theta (\ell_{r}(x', \{\hat{x}\}, \theta) + $ \\
            $\qquad \qquad \qquad \quad  \quad \quad \mathcal {L}(\hat{x}', \{y\}, \theta))]$
		\ELSE
		\STATE $\theta \gets \theta - \gamma\mathbb{E}_{x \in B}[\nabla_\theta \mathcal{L}(x', \{y\}, \theta)]$
		\ENDIF
	\ENDFOR
	\ENDFOR
	\end{algorithmic}
\end{algorithm}

\subsection{3D Object Detection Subnetwork}
Our 3D object detection model is schematically presented in Figure 2.

\paragraph{2D Information Extraction.}
The model adopts a 2D detector based on CenterNet\cite{zhou2019objects} for initial object localization and confidence scoring. This detector generates a heatmap representing coarse object locations. Alongside, it predicts the 2D offset and size for each candidate 2D bounding box. Thus $\ell_{2d}$ is shown as follow:

\begin{equation}
	\ell_{2d} = \ell_H + \ell_{O_{2d}} + \ell_{S_{2d}},
\end{equation}

\noindent where $\ell_H, \ell_{O_{2d}}, \ell_{S_{2d}}$ are Focal loss, L1 loss and GIoU loss\cite{rezatofighi2019generalized} respectively.

\paragraph{3D Information Extraction.}
Upon extracting 2D cues, our 3D object detection framework, delineated in Figure 2, intricately hones in on the Region of Interest (RoI) pertaining to the detected object. We judiciously employ the RoIAlign technique\cite{he2017mask}, facilitating a methodical cropping and resizing of the RoI features. Our model is architecturally equipped with an array of prediction sub-heads, each meticulously designed for the nuanced estimation of 3D bounding box parameters. These encompass:

A 3D offset branch, conceptualized to project the intricate 3D centroid onto the 2D feature landscape with L1 loss function $\ell_{O_{3d}}$.
A dedicated 3D size branch, aimed at extrapolating the three-dimensional parameters, specifically height, width, and length with L1 loss function $\ell_{S_{3d}}$.
An orientation prediction branch, exclusively crafted for the computation of the relative $\alpha$ rotational disposition with Multi Bin loss\cite{chen2020monopair} $\ell_{o}$.

The overall loss function of 3D information is as follow: 
\begin{equation}
	\ell_{3d} = \ell_{O_{3d}} + \ell_{S_{3d}} + \ell_{o}.
\end{equation}

\paragraph{Depth Information Extraction.}
In accordance with the methodology detailed in \cite{peng2022did}, we make an assumption that depth prediction adheres to a Laplace distribution. Thus, for each object depth $D_{ins}$ and its associated uncertainty $\sigma _{D}$, they comply with the Laplace distribution $La(D_{ins}, \sigma_{D})$. This distribution is supervised by the loss term:

\begin{equation}
	\ell_{d e p t h}=\frac{\sqrt{2}}{\sigma_D}\left|D_{ins}-d^{g t}\right|+\log \left(\sigma_D\right).
\end{equation}

\begin{table*}[ht]
	\centering
	\small
	\centering
	\renewcommand\arraystretch{1.1} 
	\setlength{\tabcolsep}{0.8mm}
	  \begin{tabular}{|l|>{\centering}p{0.9cm}>{\centering}p{0.9cm}>{\centering}p{0.9cm}|>{\centering}p{0.9cm}>{\centering}p{0.9cm}>{\centering}p{0.9cm}|>{\centering}p{0.9cm}>{\centering}p{0.9cm}>{\centering}p{0.9cm}|>{\centering}p{0.9cm}>{\centering}p{0.9cm}c|}
	  \hline
	  \multirow{2}{*}{Attack} & \multicolumn{3}{c|}{GUPNet@Car 3D} & \multicolumn{3}{c|}{MonoDLE@Car 3D} & \multicolumn{3}{c|}{MonoDDE@Car 3D} & \multicolumn{3}{c|}{DID-M3D@Car 3D} \\
  \cline{2-13}          & \cellcolor[rgb]{ .906,  .902,  .902}Easy & \cellcolor[rgb]{ .906,  .902,  .902}Mod. & \cellcolor[rgb]{ .906,  .902,  .902}Hard & \cellcolor[rgb]{ .906,  .902,  .902}Easy & \cellcolor[rgb]{ .906,  .902,  .902}Mod. & \cellcolor[rgb]{ .906,  .902,  .902}Hard & \cellcolor[rgb]{ .906,  .902,  .902}Easy & \cellcolor[rgb]{ .906,  .902,  .902}Mod. & \cellcolor[rgb]{ .906,  .902,  .902}Hard & \cellcolor[rgb]{ .906,  .902,  .902}Easy & \cellcolor[rgb]{ .906,  .902,  .902}Mod. & \cellcolor[rgb]{ .906,  .902,  .902}Hard \\
	  \hline
	  Clean & 22.704 & 16.440 & 13.671 & 17.231 & 12.262 & 10.299 & 23.410 & 16.310 & 13.310 & 23.870 & 17.110 & 13.590 \\
    \hline
    FGSM  & 7.938 & 5.493 & 4.930 & 7.673 & 4.535 & 4.239 & 8.260 & 5.762 & 4.940 & 8.503 & 6.192 & 5.111 \\
    PGD   & 3.271 & 2.229 & 2.105 & 2.959 & 2.050 & 1.393 & 3.343 & 2.398 & 2.128 & 3.934 & 3.045 & 3.003 \\
    MIFGSM & 3.455 & 2.397 & 2.128 & 2.547 & 2.104 & 1.399 & 3.308 & 3.139 & 2.894 & 3.936 & 2.433 & 2.937 \\
    IDP   & 2.115 & 1.039 & 1.011 & 1.238 & 0.855 & 0.712 & 2.389 & 0.740 & 0.932 & 2.079 & 1.368 & 1.277 \\
    \hline
	  \end{tabular}%
   \vspace{-0.8em}
   \caption{Performance comparison of the most popular monocular 3D object detection methods under clean images and various adversarial attacks.}
	\label{tab:addlabel}%
\end{table*}%
  
\begin{table*}[ht]
	\centering
	\small
	\centering
	\renewcommand\arraystretch{1.1} 
	\setlength{\tabcolsep}{0.8mm}  
	\begin{tabular}{|l|>{\centering}p{0.9cm}>{\centering}p{0.9cm}>{\centering}p{0.9cm}|>{\centering}p{0.9cm}>{\centering}p{0.9cm}>{\centering}p{0.9cm}|>{\centering}p{0.9cm}>{\centering}p{0.9cm}>{\centering}p{0.9cm}|>{\centering}p{0.9cm}>{\centering}p{0.9cm}>{\centering}p{0.9cm}|>{\centering}p{0.9cm}>{\centering}p{0.9cm}c|}
		\hline
	  \multirow{2}{*}{Attack} & \multicolumn{3}{c|}{GUPNet@Car 3D} & \multicolumn{3}{c|}{MonoDLE@Car 3D} & \multicolumn{3}{c|}{MonoDDE@Car 3D} & \multicolumn{3}{c|}{DID-M3D@Car 3D} & \multicolumn{3}{c|}{Ours@Car 3D} \\
  \cline{2-16}          & \cellcolor[rgb]{ .906,  .902,  .902}Easy & \cellcolor[rgb]{ .906,  .902,  .902}Mod. & \cellcolor[rgb]{ .906,  .902,  .902}Hard & \cellcolor[rgb]{ .906,  .902,  .902}Easy & \cellcolor[rgb]{ .906,  .902,  .902}Mod. & \cellcolor[rgb]{ .906,  .902,  .902}Hard & \cellcolor[rgb]{ .906,  .902,  .902}Easy & \cellcolor[rgb]{ .906,  .902,  .902}Mod. & \cellcolor[rgb]{ .906,  .902,  .902}Hard & \cellcolor[rgb]{ .906,  .902,  .902}Easy & \cellcolor[rgb]{ .906,  .902,  .902}Mod. & \cellcolor[rgb]{ .906,  .902,  .902}Hard & \cellcolor[rgb]{ .906,  .902,  .902}Easy & \cellcolor[rgb]{ .906,  .902,  .902}Mod. & \cellcolor[rgb]{ .906,  .902,  .902}Hard \\
	  \hline
	  Clean & 18.540 & 13.599 & 10.217 & 14.790 & 10.060 & 7.830 & 19.010 & 13.430 & 10.370 & 18.770 & 13.160 & 9.810 & 19.700 & 14.910 & 12.016 \\
	  \hline
	  FGSM  & 14.614 & 10.482 & 7.996 & 12.610 & 8.586 & 6.615 & 13.727 & 10.164 & 8.600 & 13.624 & 9.427 & 7.461 & 16.879 & 12.362 & 10.021 \\
	  PGD   & 11.731 & 8.213 & 6.581 & 11.651 & 8.019 & 3.917 & 10.383 & 8.858 & 5.457 & 10.957 & 6.240 & 6.673 & 13.999 & 9.266 & 7.535 \\
	  MIFGSM & 11.474 & 8.283 & 4.039 & 9.397 & 8.449 & 5.434 & 9.345 & 6.735 & 5.482 & 9.406 & 6.735 & 4.570 & 12.956 & 9.727 & 6.521 \\
	  IDP   & 6.935 & 4.934 & 3.190 & 7.649 & 4.756 & 5.271 & 6.970 & 5.162 & 3.665 & 7.536 & 5.965 & 2.546 & 10.562 & 7.998 & 6.210 \\
	  \hline
	  \end{tabular}%
   \vspace{-0.8em}
   \caption{Comparison of different methods after undergoing adversarial training based on IDP, evaluating their performance under both clean conditions and various attacks on the KITTI 3D detection dataset on ${AP}_{3D}$ of car category. Quantitative results prove that our method achieves state-of-the-art performance.}
	\label{tab:addlabel}%
\end{table*}%

\section{Experiments}
In this section, we first present the experimental setup and implementation details. Subsequently, we conduct experiments on the KITTI 3D dataset to validate the effectiveness of our proposed adversarial attacks and the impact of different loss functions on the robustness of 3D object detection models. Furthermore, we demonstrate that our proposed defense method outperforms other adversarial training approaches.

\subsection{Implementation Details}

\begin{figure}[ht]
	\centering
	\includegraphics[width=0.48\textwidth,height=0.14\textheight]{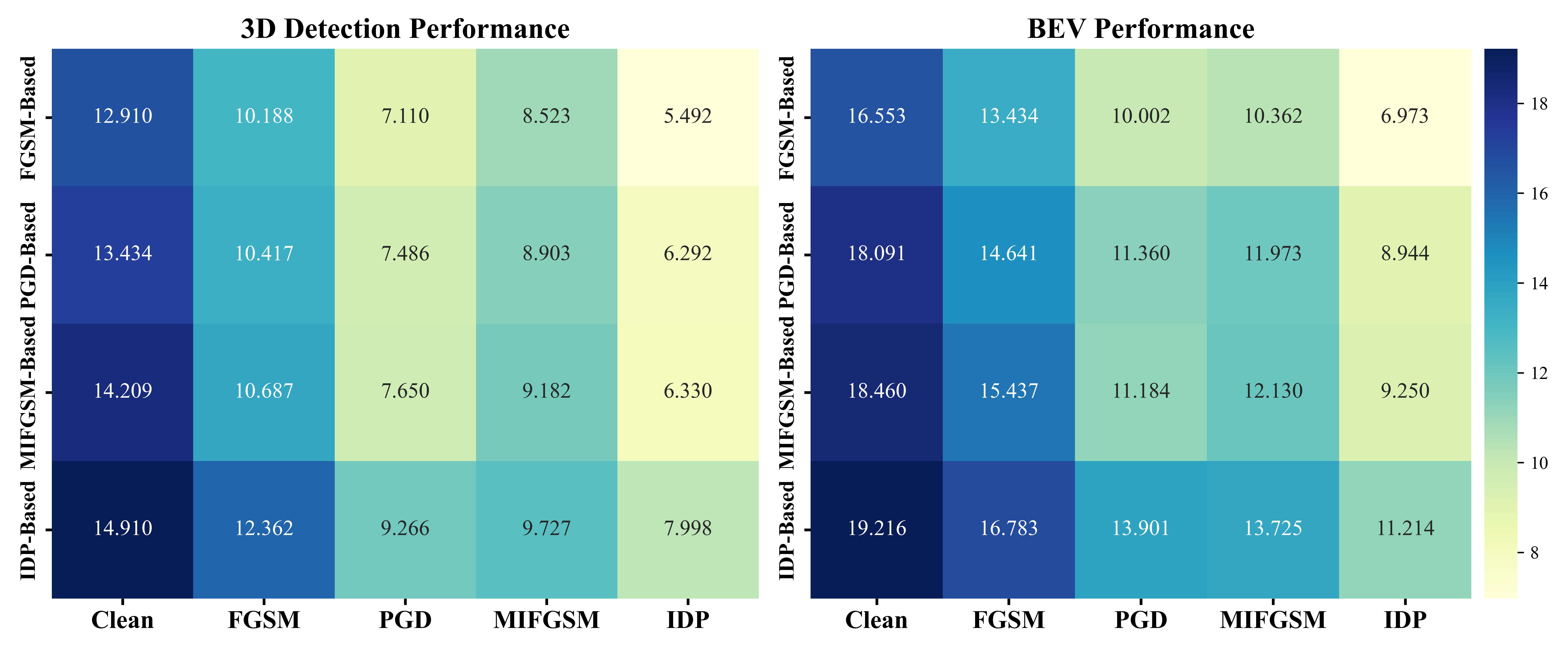}
	\vspace{-0.8em}
	\caption{Comparison of our method's 3D detection performance under adversarial training based on FGSM, PGD, MIFGSM, and IDP attacks, evaluated in both clean conditions and various attack scenarios.}
\end{figure}

We carry out our experiments on 4 NVIDIA RTX TITAN XP GPUs, utilizing a batch size of 16. Our implementation is based on the PyTorch framework \cite{paszke2019pytorch}. The network is trained for 140 epochs, following the Hierarchical Task Learning (HTL) \cite{lu2021geometry} training strategy. We employ the Adam optimizer with an initial learning rate of 1e-5. A linear warm-up strategy is used to increase the learning rate to 1e-3 within the first 5 epochs. The learning rate then decays at epoch 50 and 80 at a rate of 0.1. For multi-bin orientation $\theta$, we set k to 12. The backbone and head architecture adheres to the design presented in \cite{lu2021geometry}. The input image is resized to a resolution of 1280 × 384, with pixel values ranging between [0, 255]. The pixel intensities are subsequently shifted according to the mean pixel intensity of the entire dataset. For adversarial training, we generate adversarial examples using a budget $\epsilon$ = 3 as the inputs, set the PGD step size to 2 and limit it to 3 steps.

\subsection{Dataset and Metrics}

In accordance with the experimental setup commonly adopted in prior works \cite{brazil2019m3d,ding2020learning,liu2021autoshape}, we conduct our experiments on the widely used KITTI 3D detection dataset \cite{geiger2012we}. The KITTI dataset comprises 7,481 training samples and 7,518 testing samples, with the training sample labels being publicly available while the testing sample labels remain undisclosed on the KITTI website, reserved solely for online evaluation and ranking purposes. To facilitate ablation studies, previous research has further partitioned the 7,481 samples into a new training set of 3,712 samples and a validation set of 3,769 samples. This data split \cite{chen20173d} has been widely employed in numerous previous works. Moreover, KITTI categorizes objects into easy, moderate, and hard levels based on the 2D bounding box height (correlated to depth), occlusion, and truncation levels. For evaluation metrics, we utilize the recommended AP40 metric \cite{simonelli2019disentangling} for the two primary tasks, namely 3D and bird's-eye-view (BEV) detection.

\begin{figure*}[t]
	\centering
	\includegraphics[scale=0.61]{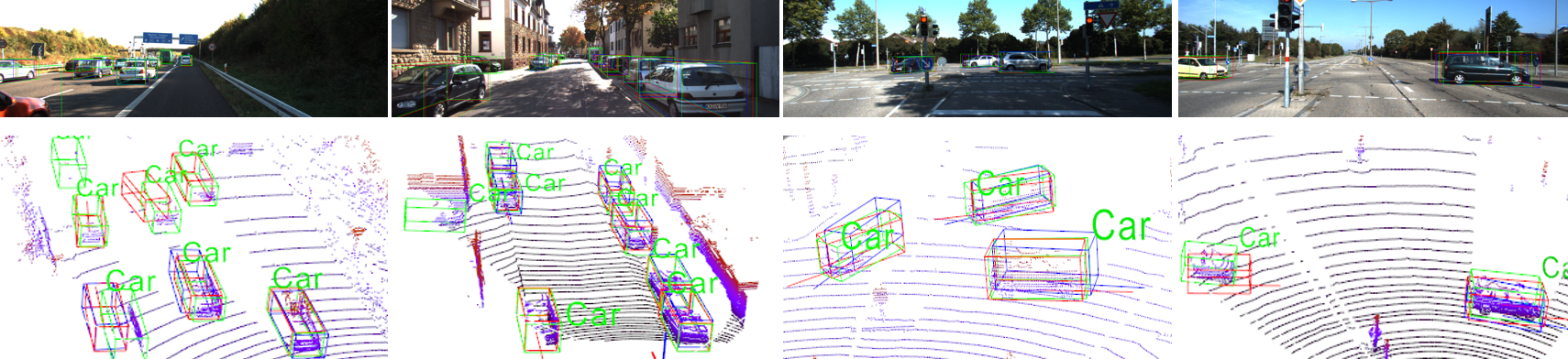}
	\vspace{-0.6em}
	\caption{A comparative analysis of the quality results between our method and DID-M3D under IDP attacks. Our approach demonstrates superior robustness, exhibiting fewer instances of missed targets, and consistently produces more precise 3D bounding boxes.}
\end{figure*}

\begin{figure*}[t]
	\centering
	\includegraphics[scale=0.53]{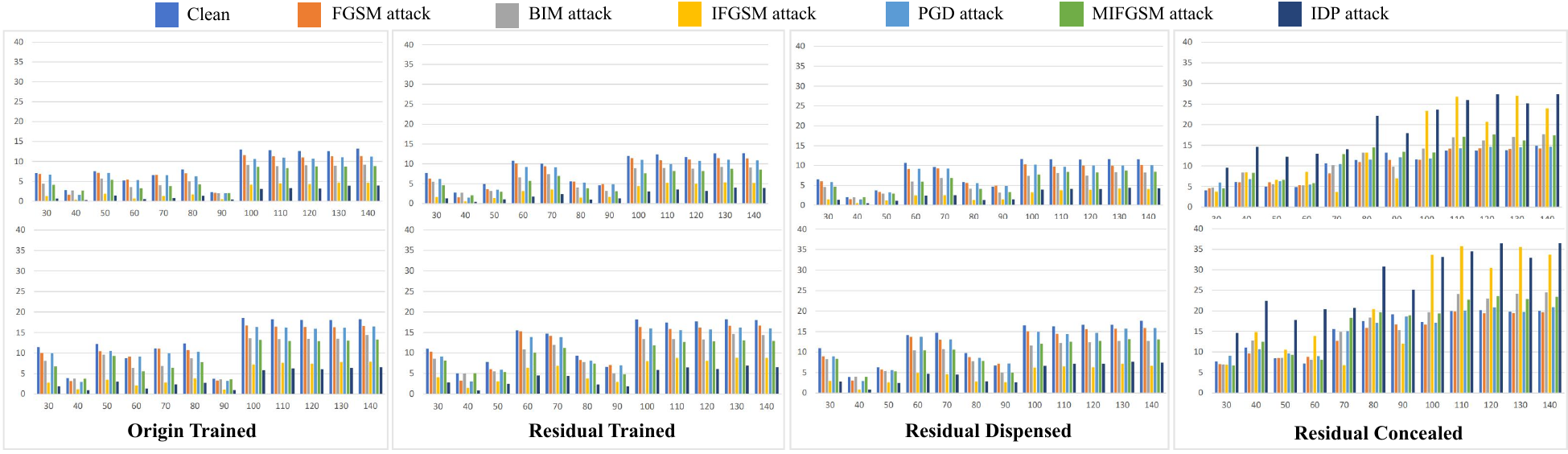}
	\caption{A comparison of 3D object detection results ${Car}_{3D}$ (rop row) and${Car}_{BEV}$ (bottom row) under clean images and six attack types (FGSM\cite{goodfellow2014explaining}, BIM\cite{kurakin2016adversarial}, IFGSM\cite{kurakin2018adversarial}, PGD\cite{madry2017towards}, MIFGSM\cite{dong2018boosting} and IDP) during epochs 30-140. The grouped bar chart consists of eight sets of bars representing four scenarios.}
\end{figure*}

\subsection{Evaluation Results of 3D Detection Models}
We conducted a comparative analysis of four prominent monocular 3D object detection methods: GUPNet\cite{lu2021geometry}, MonoDLE\cite{ma2021delving}, MonoDDE\cite{li2022diversity}, and DID-M3D\cite{peng2022did}. Their performance was evaluated on clean images as well as under various adversarial attacks, including FGSM\cite{goodfellow2014explaining}, PGD\cite{madry2017towards}, MIFGSM\cite{dong2018boosting}, and IDP. As illustrated in Table~1, all detectors exhibited a significant degradation in performance under adversarial conditions. Notably, under the IDP attack, the performance of DID-M3D, which originally achieved scores of 23.870\%, 17.110\%, and 13.590\% on clean images, plummeted to 2.079\%, 1.368\%, and 1.277\%, respectively. This stark decline underscores the vulnerability of state-of-the-art monocular 3D object detection methods in the face of adversarial attacks.

\subsection{Performance of Adversarial Training}
In this section, we conduct a comprehensive performance evaluation of various methods on the KITTI 3D Benchmark under clean image, FGSM, PGD, MIFGSM, and IDP attack scenarios. The methods under comparison include GPUNet, MonoDLE, MonoDDE, DID-M3D trained with IDP attack, as well as our proposed uncertainty-based residual learning approaches.

Table 2 presents the 3D $AP_{R40}$ results for the car category. Our method achieves detection accuracies of 19.700 , 14.910 , and 12.016  for the Easy, Moderate, and Hard setting, respectively, on clean data. This represents an improvement of 3.63\%, 11.03\%, and 15.88\% over the second-best method, MonoDDE, which achieved performance of 19.010, 13.430 , and 10.370 , respectively. Under adversarial attacks such as FGSM, PGD, and MIFGSM, our method consistently outperforms the state-of-the-art. Notably, under the IDP attack, our method achieves scores of 10.562, 7.998, and 6.210, marking a significant improvement of 40.12\%, 34.07\%, and 143.90\% over DID-M3D's performance of 7.536, 5.965, and 2.546, respectively.

As depicted in Figure 3, we further contrast the adversarial robustness of our method post adversarial training under various attacks. Empirical evidence suggests that training under the IDP attack can enhance the model's 3D detection  and BEV performance. 

In summary, our proposed methods exhibit consistently outstanding performance in both 3D and BEV $AP_{R40}$ metrics.

\begin{table}[t]
	\centering
 \footnotesize
	\renewcommand{\arraystretch}{0.9}
	\setlength{\tabcolsep}{1.1mm}{
	\begin{tabular}{l|c|rrr|rrr}
	  \toprule[1pt]
	  \multirow{2}{*}{ReL} & \multicolumn{1}{c|}{\multirow{2}{*}{UnB}} & \multicolumn{3}{c|}{3D $AP_{R40}$} & \multicolumn{3}{c}{BEV $AP_{R40}$} \\
  \cline{3-8}          &     & \multicolumn{1}{c}{\cellcolor[rgb]{ .906,  .902,  .902}Easy} & \multicolumn{1}{c}{\cellcolor[rgb]{ .906,  .902,  .902}Mod.} & \multicolumn{1}{c|}{\cellcolor[rgb]{ .906,  .902,  .902}Hard} & \multicolumn{1}{c}{\cellcolor[rgb]{ .906,  .902,  .902}Easy} & \multicolumn{1}{c}{\cellcolor[rgb]{ .906,  .902,  .902}Mod.} & \multicolumn{1}{c}{\cellcolor[rgb]{ .906,  .902,  .902}Hard} \\
	  \midrule
	  -     & -        & 6.111 & 3.876 & 3.015 & 10.223 & 6.572 & 5.128 \\
	  \checkmark & -          & 6.455 & 3.997 & 3.157 & 10.759 & 6.813 & 5.472 \\
	  -     & \checkmark  & 6.997 & 3.935 & 3.348 & 11.659 & 7.245 & 5.941 \\
	  \checkmark & \checkmark & 10.562 & 7.998 & 6.210 & 14.328 & 11.214 & 9.113\\
	\bottomrule[1pt]
	\end{tabular}%
	}
 \vspace{-0.8em}
 \caption{Ablation study for the effectiveness of the components of our approach, our experiments are reported on the KITTI 3D dataset under IDP attack.}
	\label{tab:addlabel}%
\end{table}%

\begin{table}[t]
	\centering
	\scriptsize
	\renewcommand\arraystretch{1} 
	\setlength{\tabcolsep}{0.67mm}
	\begin{tabular}{|c|c|ccc|ccc|}
	  \hline
	  \multirow{2}{*}{Att.} & \multirow{2}{*}{Denoiser} & \multicolumn{3}{c|}{3D Det} & \multicolumn{3}{c|}{BEV} \\
  \cline{3-8}          &       & \cellcolor[rgb]{ .906,  .902,  .902}Easy & \cellcolor[rgb]{ .906,  .902,  .902}Mod. & \cellcolor[rgb]{ .906,  .902,  .902}Hard & \cellcolor[rgb]{ .906,  .902,  .902}Easy & \cellcolor[rgb]{ .906,  .902,  .902}Mod. & \cellcolor[rgb]{ .906,  .902,  .902}Hard \\
	  \hline
	  \multirow{3}{*}{\begin{sideways}Clean\end{sideways}} & Noise2Noise & 19.108  & 14.910  & 10.176  & 23.833  & 18.178  & 13.932  \\
			& Noise2Void & 19.289  & 14.627  & 12.107  & 24.692  & 18.576  & 15.855  \\
			& ours  & 19.700  & 14.910  & 12.016  & 25.127  & 19.216  & 15.334  \\
	  \hline
	  \multirow{3}{*}{\begin{sideways}FGSM\end{sideways}} & Noise2Noise & 15.777  & 11.830  & 8.993  & 22.007  & 15.461  & 12.744  \\
			& Noise2Void & 16.429  & 12.698  & 8.378  & 21.649  & 16.937  & 12.217  \\
			& ours  & 16.879  & 12.362  & 10.021  & 22.787  & 16.783  & 13.219  \\
	  \hline
	  \multirow{3}{*}{\begin{sideways}PGD\end{sideways}} & Noise2Noise & 12.525  & 7.985  & 7.010  & 18.302  & 12.353  & 10.359  \\
			& Noise2Void & 12.883  & 9.712  & 6.606  & 19.714  & 12.948  & 11.715  \\
			& ours  & 13.999  & 9.266  & 7.535  & 19.554  & 13.901  & 11.238  \\
	  \hline
	  \multirow{3}{*}{\begin{sideways}MIFGSM\end{sideways}} & Noise2Noise & 11.149  & 7.977  & 5.893  & 19.458  & 12.043  & 11.008  \\
			& Noise2Void & 12.015  & 9.084  & 6.525  & 18.790  & 13.386  & 10.740  \\
			& ours  & 12.956  & 9.727  & 6.521  & 19.963  & 13.725  & 11.759  \\
	  \hline
	  \multirow{3}{*}{\begin{sideways}IDP\end{sideways}} & Noise2Noise & 9.262  & 6.460  & 4.217  & 16.737  & 10.347  & 9.059  \\
			& Noise2Void & 10.205  & 7.860  & 5.493  & 18.958  & 10.722  & 8.890  \\
			& ours  & 10.562  & 7.998  & 6.210  & 18.632  & 11.214  & 9.994  \\
	  \hline
	  \end{tabular}%
   \vspace{-0.8em}
   \caption{Performance comparison of 3D object detection under various attacks, utilizing different denoisers including Noise2Noise, Noise2Void, and our proposed method.}

	\label{tab:addlabel}%
\end{table}%

\begin{figure}[ht]
	\centering
    \tiny
    \renewcommand\arraystretch{0.8} 
	\setlength{\tabcolsep}{0.6pt}
	\begin{tabular}{cc}	
		\includegraphics[width=0.46\textwidth,height=0.16\textheight]{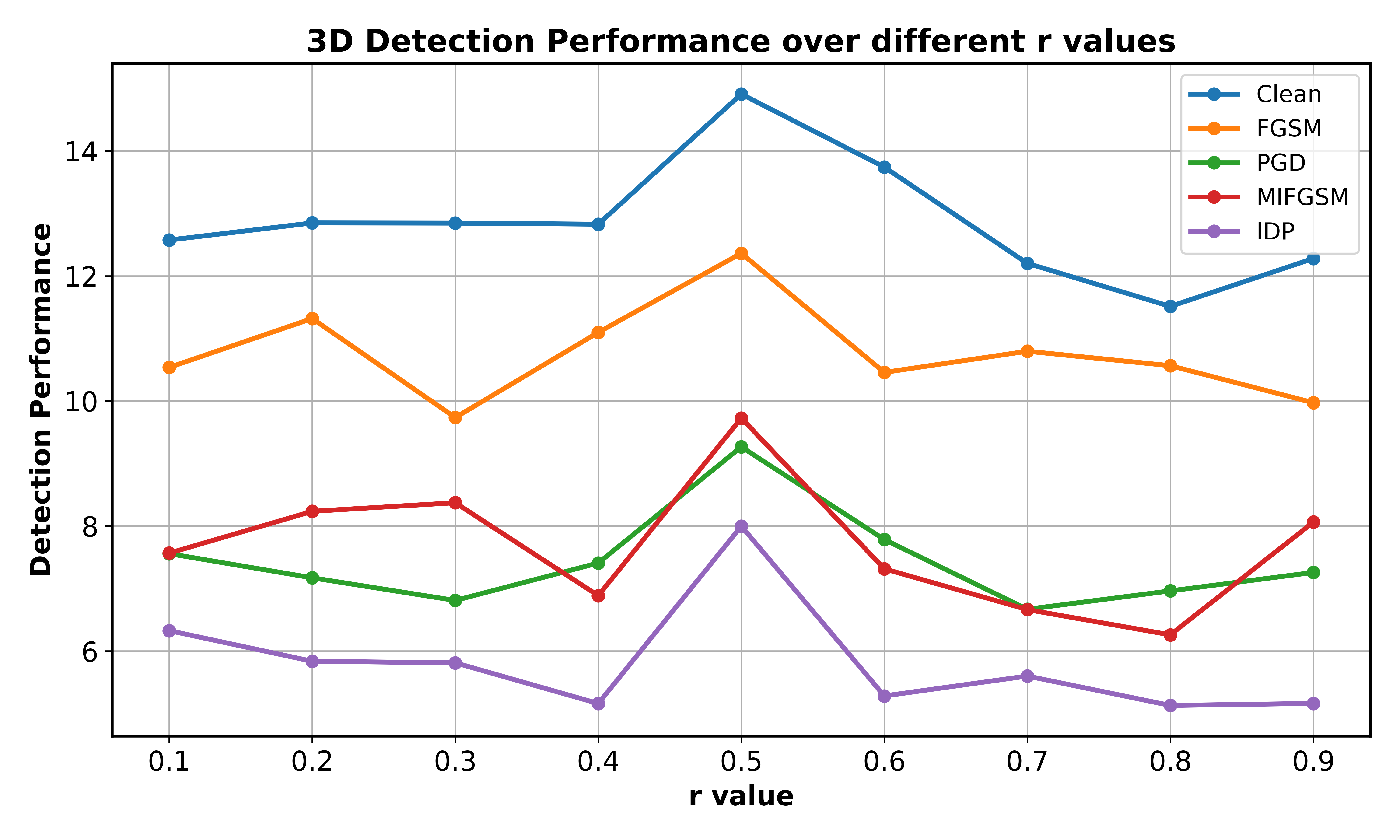} \\
	\includegraphics[width=0.46\textwidth,height=0.16\textheight]{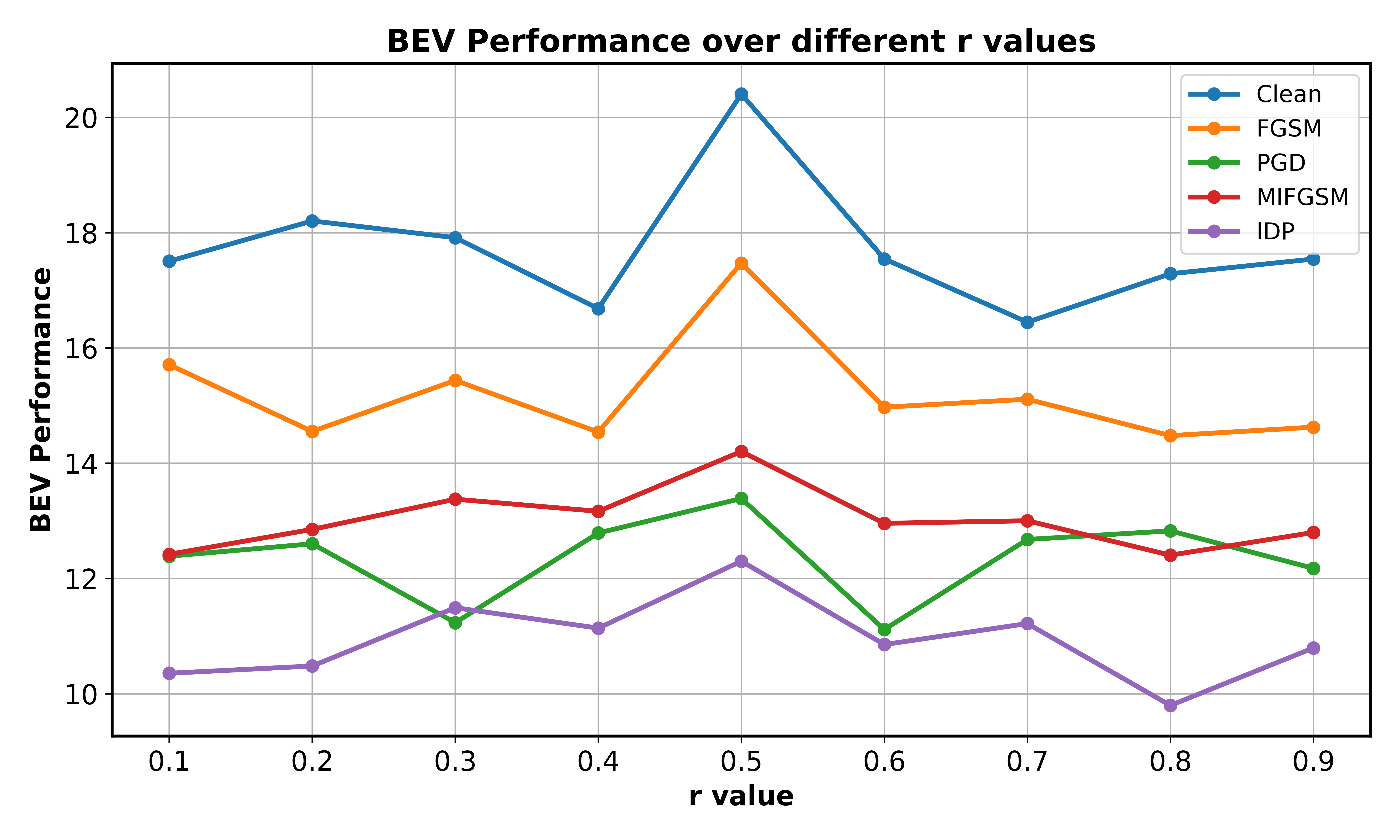}\\
        \end{tabular}\vspace{-1em}
	\caption{Ablation study illustrating the performance variation of our model with respect to the threshold value of $r$ ranging from 0.1 to 0.9. The evaluation encompasses both 3D detection and BEV metrics under clean conditions as well as under FGSM, PGD, MIFGSM, and IDP adversarial attacks.}
	\label{fig:exp6}
\end{figure}

\subsection{Ablation Study}
In this section, we conduct comprehensive and rigorous experiments to meticulously assess the robustness and efficacy of our proposed approach, delving deeply into the nuanced impact of each integral component within our framework.

\paragraph{Ablation Study on Key Components.}
Table 3 provides an ablation analysis of our two pivotal components: residual learning module (ReL) and uncertainty-based module (UnB) . Empirical results suggest that while each component contributes individually, it is their combined utilization that yields the optimal performance.

\paragraph{Analysis of Residual Learning.}
As illustrated in Figure 5, we delve into various configurations related to the employment of residual learning. Specifically, we examine: Origin Trained (scenarios without any use of residual learning), Residual Trained (our model incorporating residual learning while the attacker is aware of it), Residual Dispensed (our model abstaining from residual learning but the attacker considering it) and Residual Concealed (our model leveraging residual learning, yet the attacker remains oblivious to it). The detection accuracy and BEV precision across different epochs for these configurations are presented. Notably, there's a discernible trend of improvement from one scenario to the next, underscoring the remarkable efficacy of uncertainty in countering adversarial attacks. However, given the challenges in actualizing scenarios 3 and 4, our approach predominantly adopts the second scenario.

\paragraph{Experiments with Various Denoisers.}
Table 4 showcases the performance of our adversarial training approach based on uncertainty across different denoisers, namely Noise2Noise, Noise2Void, and our proposed method. The 3D detection results under various attacks indicate that while all denoisers exhibit robustness, our method, in synergy with uncertainty, achieves the most commendable performance.

\paragraph{Impact of Uncertainty Threshold.}
Figure 6 elucidates the effect of varying the threshold \( r \) in the uncertainty-based residual learning on the performance of 3D detection and BEV under the moderate setting. As depicted, the model exhibits consistent robustness across thresholds ranging from 0.1 to 0.9. Notably, optimal performance on both metrics is observed when \( r < 0.5 \).

\subsection{Qualitative results}

To provide a more intuitive demonstration of the superiority of our approach, we visualize the detection results of our model and DID-M3D, both subjected to the same adversarial training based on IDP, as shown in Figure 4. On highways with high traffic volume and crowded vehicles, DID-M3D is prone to missing obstructed vehicles while our method can almost identify all vehicles and the location information is more accurate. On the path, our method also detects the object position more accurately, with little difference from the correct result. At intersections and pedestrian areas, for vehicles in motion, the recognition results of other methods show significant deviation from the correct results, and sometimes there is a problem of incomplete recognition. We have more comprehensive and accurate detection results, which almost overlap with the correct results. In situations involving occlusion and long distances, the competing method tends to miss objects while our results are closer to the ground truth, highlighting the enhanced accuracy of our approach. Therefore, our method outperforms current adversarial training techniques significantly.

\section{Conclusion}
In this work, we introduce the IDP attack, an adversarial attack meticulously crafted for monocular 3D object detection. This iterative attack strategically targets both 2D and 3D depth information, aiming to severely compromise the detection capabilities of contemporary models. To counteract such potent threats, we propose DART3D, an network fortified with uncertainty-based robustness. Central to DART3D's design is a residual learning module, which harnesses the power of uncertainty-driven residual learning for denoising. This module adeptly extracts salient features from adversarial samples, effectively mitigating the perturbations introduced during attacks. Comprehensive evaluations on the KITTI 3D Benchmark underscore the prowess of our approach.
\bibliography{aaai23}

\end{document}